# Solution-Guided Multi-Point Constructive Search for Job Shop Scheduling

**J. Christopher Beck**                                            jcb@mie.utoronto.ca
*Department of Mechanical & Industrial Engineering*
*University of Toronto, Canada*

## Abstract

Solution-Guided Multi-Point Constructive Search (SGMPCS) is a novel constructive search technique that performs a series of resource-limited tree searches where each search begins either from an empty solution (as in randomized restart) or from a solution that has been encountered during the search. A small number of these "elite" solutions is maintained during the search. We introduce the technique and perform three sets of experiments on the job shop scheduling problem. First, a systematic, fully crossed study of SGMPCS is carried out to evaluate the performance impact of various parameter settings. Second, we inquire into the diversity of the elite solution set, showing, contrary to expectations, that a *less* diverse set leads to stronger performance. Finally, we compare the best parameter setting of SGMPCS from the first two experiments to chronological backtracking, limited discrepancy search, randomized restart, and a sophisticated tabu search algorithm on a set of well-known benchmark problems. Results demonstrate that SGMPCS is significantly better than the other constructive techniques tested, though lags behind the tabu search.

## 1. Introduction

A number of metaheuristic and evolutionary approaches to optimization can be described as being "solution-guided, multi-point" searches. For example, in genetic and mimetic algorithms, a population of solutions is maintained and used as a basis for search. Each new generation is created by combining aspects of the current generation: search is therefore guided by existing solutions. As the population contains a number of individual solutions, the search makes use of multiple points in the search space. Traditional single-point metaheuristics, such as tabu search, have been augmented in a similar way. The TSAB tabu search (Nowicki & Smutnicki, 1996) maintains an elite pool consisting of a small number of the best solutions found so far during the search. Whenever the basic search reaches a threshold number of moves without finding a new best solution, search is restarted from one of the elite solutions. Again, the higher-level search is guided by multiple existing solutions, though the guidance is somewhat different than in genetic algorithms.

Solution-Guided Multi-Point Constructive Search (SGMPCS)[1] is a framework designed to allow constructive search to be guided by multiple existing (suboptimal) solutions to a problem instance. As with randomized restart techniques (Gomes, Selman, & Kautz, 1998), the framework consists of a series of tree searches restricted by some resource limit, typically a maximum number of fails. When the resource limit is reached, search restarts. The difference with randomized restart is that SGMPCS keeps track of a small set of "elite solutions": the best solutions it has found. When search is restarted, it starts from an empty solution, as in randomized restart, or from one of the elite solutions.

---

1. In previous conference and workshop publications, SGMPCS is referred to simply as Multi-Point Constructive Search (Beck, 2006; Heckman & Beck, 2006; Beck, 2005a, 2005b). Empirical evidence of the importance of solution guidance motivated this change to a name more reflective of the important differences between this work and existing tree search techniques.

In this paper, we undertake the first fully crossed systematic empirical study of SGMPCS. In particular, in Section 3 we investigate the different parameter settings and their impact on search performance for the makespan-minimization variant of the job shop scheduling problem. Results indicate that guidance with elite solutions contributes significantly to algorithm performance but, somewhat unexpectedly, that smaller elite set size results in better performance. Indeed, an elite set size of one showed the best performance. This result motivates subsequent experimentation on the diversity of the elite set in Section 4. We show, again contrary to expectation but consistent with an elite set size of one, that the *less* diverse the elite set, the stronger the performance. As discussed in-depth in Section 6, these two sets of experiments call into question the extent to which the exploitation of multiple points in the search space is important for the performance of SGMPCS. A final experiment (Section 5) compares the best parameter settings found in the first two experiments with chronological backtracking, limited discrepancy search (Harvey, 1995), randomized restart, and a state-of-the-art tabu search (Watson, Howe, & Whitley, 2006) on a set of well-known benchmarks. These results show that SGMPCS significantly out-performs the other constructive search methods but does not perform as well as the tabu search.

The contributions of this paper are as follows:

1. The introduction and systematic experimental evaluation of Solution-Guided Multi-Point Constructive Search (SGMPCS).

2. The investigation of the importance of the diversity of the elite set to the performance of SGMPCS.

3. The demonstration that SGMPCS significantly out-performs chronological backtracking, limited discrepancy search, and randomized restart on a benchmark set of job shop scheduling problems.

## 2. Solution-Guided Multi-Point Constructive Search

Pseudocode for the basic Solution-Guided Multi-Point Constructive Search algorithm is shown in Algorithm 1. The algorithm initializes a set, $e$, of elite solutions and then enters a while-loop. In each iteration, with probability $p$, search is started from an empty solution (line 6) or from a randomly selected elite solution (line 12). In the former case, if the best solution found during the search, $s$, is better than the worst elite solution, $s$ replaces the worst elite solution. In the latter case, $s$ replaces the starting elite solution, $r$, if $s$ is better than $r$. Each individual search is limited by a maximum number of fails that can be incurred. When an optimal solution is found and proved or when some overall bound on the computational resources (e.g., CPU time, number of fails) is reached, the best elite solution is returned.

The elite solutions can be initialized by any search technique. In this paper, we use 50 independent runs of the same randomized texture-based heuristic that is employed in the

**SGMPCS**():

**1** initialize elite solution set $e$
**2** **while** *termination criteria unmet* **do**
**3**    **if** $rand[0,1) < p$ **then**
**4**       set upper bound on cost function
**5**       set fail limit, $l$
**6**       $s := \text{search}(\emptyset, l)$
**7**       **if** $s \neq \emptyset$ *and s is better than worst(e)* **then**
**8**          replace worst($e$) with $s$
         **end**
      **else**
**9**       $r := $ randomly chosen element of $e$
**10**      set upper bound on cost function
**11**      set fail limit, $l$
**12**      $s := \text{search}(r, l)$
**13**      **if** $s \neq \emptyset$ *and s is better than r* **then**
**14**         replace $r$ with $s$
         **end**
      **end**
   **end**
**15** return best($e$)

**Algorithm 1**: SGMPCS: Solution-Guided Multi-Point Constructive Search

main search (see Section 3.2). No backtracking is done and no upper bound is placed on the cost function. Without an upper bound, each run will find a solution, though probably one of quite low quality. From this initial set of 50 solutions, the $|e|$ best solutions are inserted into the elite set. The primary goal for the initialization is to quickly populate the elite set. Previous work (Beck, 2006) has shown that while spending more effort in each run to find good starting solutions (e.g., via backtracking search) does not significantly improve overall performance, the number of runs does have an impact. When the variance in quality among the initial solutions is high, the best starting solution of a large elite set will be much better than that of a small elite set. This difference alone was sufficient to skew experiments that measured the impact of different elite set sizes on overall performance. To mitigate this effect we generate a fixed number of elite solution candidates (i.e., 50) and then choose the $|e|$ best. An interesting direction for future work is to adaptively determine the best time to transition from the elite pool generation to the main search.

### 2.1 Search

In lines 6 and 12 the search($r, l$) function is a standard tree search with some randomization, limited by the number of fails, $l$, and, when $r \neq \emptyset$, guided by solution $r$. The search function returns the best solution found, if any, and an indication as to whether the search space has been exhausted. Given a large enough fail limit, an individual search can completely search the space. Therefore, the completeness of this approach depends on the policy for setting and increasing the fail limit. As we will see in Experiment 3 (Section 5), SGMPCS is able to find optimal solutions and prove their optimality. We place no other restrictions on the search, allowing any tree traversal technique to be used. In particular, we experiment with both chronological backtracking and limited discrepancy search (Harvey, 1995).

When $r \neq \emptyset$, the search is guided by the *reference* solution, $r$. The guiding solution is simply used as the value ordering heuristic: we search using any (randomized) variable ordering heuristic and by specifying that the value assigned to a variable is the one in the reference solution, provided it is still in the domain of the variable.

A search tree is created by asserting a series of choice points of the form: $\langle V_i = x \rangle \vee \langle V_i \neq x \rangle$, where $V_i$ is a variable and $x$ is the value assigned to $V_i$. Given the importance of variable ordering heuristics in constructive search, we expect that the order of these choice points will have an impact on search performance. SGMPCS can, therefore, use any variable ordering heuristic to choose the next variable to assign. The choice point is formed using the value assigned in the reference solution or, if the value in the reference solution is inconsistent, a heuristically chosen value. More formally, let a reference solution, $r$, be a set of variable assignments, $\{\langle V_1 = x_1 \rangle, \langle V_2 = x_2 \rangle, \ldots, \langle V_m = x_m \rangle\}, m \leq n$, where $n$ is the number of variables. The variable ordering heuristic has complete freedom to choose a variable, $V_i$, to be assigned. If $x_i \in dom(V_i)$, where $\langle V_i = x_i \rangle \in r$, the choice point is made with $x = x_i$. Otherwise, if $x_i \notin dom(V_i)$, any value ordering heuristic can be used to choose $x \in dom(V_i)$.

We need to account for the possibility that $x_i \notin dom(V_i)$ because the reference solution is not necessarily a valid solution later in the SGMPCS search process. To take a simple example, if the reference solution has a cost of 100 and we constrain the search to find a better solution, we will not reach the reference solution. Rather, via constraint propagation, we will reach a dead-end or different solution.

This technique for starting constructive search from a reference solution is quite general. Existing high-performance variable ordering heuristics can be exploited and, by addressing the case of $x_i \notin dom(V_i)$, we make no assumptions about changes to the constraint model that may have been made after the reference solution was originally found. In particular, this means that an elite solution could be the solution to a relaxation of the full problem.

## 2.2 Setting the Bounds on the Cost Function

Before each individual search (lines 6 and 12), we place an upper bound on the cost function. The bound has an impact on the set of solutions and, therefore, on the solutions that may enter the elite set. Intuitions from constructive search and metaheuristics differ on the appropriate choice of an upper bound. In standard tree search for optimization with a discrete cost function, the usual approach is to use $c^* - 1$ as the upper bound, where $c^*$ is the best solution found so far. Using a higher bound would only expand the search space without providing any heuristic benefit. In contrast, in a standard metaheuristic approach, search is not usually restricted by enforcing an upper bound on the cost of acceptable states: the search is allowed to travel through worse states in order to (hopefully) find better ones. As a consequence, it is common to replace an elite solution when a better, but not necessarily best-known, solution is found. Since the elite solutions are used to heuristically guide search, even solutions which are not the best-known can provide heuristic guidance.

These two perspectives result in two policies:

1. *Global Bound*: Always set the upper bound on the search cost to $c^* - 1$.

2. *Local Bound*: When starting from an empty solution, set the upper bound to be equal to one less than the cost of the worst elite solution. When starting from an elite solution, set the upper bound to be one less than the cost of the starting solution.

In constraint programming, *back-propagation* is the extent to which placing a bound on the cost function results in domain reductions for decision variables. Previous experiments with SGMPCS on optimization problems with strong back-propagation (such as job shop scheduling with the objective of minimizing makespan) show that the global bound policy is superior (Beck, 2006). For problems with weaker back-propagation and for satisfaction problems (where there is no back-propagation), the local bound approach performs better (Beck, 2006; Heckman & Beck, 2006). Based on these results, we use the global bound policy here.

### 2.3 Related Work

SGMPCS is most directly inspired by the TSAB tabu search algorithm (Nowicki & Smutnicki, 1996) noted above. In TSAB, an elite pool consisting of a small number of the best solutions found is maintained during the search. Whenever the basic tabu search stagnates, that is, when it reaches a threshold number of moves without finding a new best solution, search is restarted from one of the elite solutions. The tabu list is modified so that when search is restarted, it will follow a different search path. This is the basic mechanism, adapted for constructive search, that is used in SGMPCS. For a number of years, TSAB was the state-of-the-art algorithm for job shop scheduling problems. It has recently been over-taken by $i$-TSAB, an algorithm based on TSAB that makes a more sophisticated use of the elite pool (Nowicki & Smutnicki, 2005). For an in-depth analysis of $i$-TSAB see the work of Watson, Howe, and Whitley (2006).

SGMPCS performs a series of resource-limited tree searches. It is clear that such behaviour is related to the extensive work on randomized restart (Gomes et al., 1998; Horvitz, Ruan, Gomes, Kautz, Selman, & Chickering, 2001; Kautz, Horvitz, Ruan, Gomes, & Selman, 2002; Gomes, Fernández, Selman, & Bessière, 2005; Hulubei & O'Sullivan, 2006). Indeed, setting $p$, the probability of searching from an empty solution, to 1 results in a randomized restart technique. It has been observed that search effort for chronological backtracking and a given variable ordering forms a "heavy-tailed" distribution. Intuitively, this means that a randomly chosen variable ordering has a non-trivial chance of resulting in either a small or a large cost to find a solution to a problem instance. If no solution is found after some threshold amount of effort, it is beneficial to restart search with a different variable ordering as the new ordering has a non-trivial probability of quickly leading to a solution.

There are a number of other techniques that make use of randomized or heuristic backtracking (Prestwich, 2002; Jussien & Lhomme, 2002; Dilkina, Duan, & Havens, 2005) to form a hybrid of local search and tree search and allow an exploration of the search space that is not as constrained as standard tree search. These approaches differ from SGMPCS at the fundamental level: they do not use (multiple) existing solutions to guide search.

## 3. Experiment 1: Parameter Settings

The primary purpose of this experiment is to understand the impact of the different parameter settings on the performance of SGMPCS algorithms. We present a fully crossed experiment to evaluate the impact of varying the parameters of SGMPCS.

### 3.1 SGMPCS Parameters

**Elite Set Size**  The number of elite solutions that are maintained during the search is a key parameter controlling the extent to which multiple points in the search space are exploited by SGMPCS. While there does not seem to have been significant experimentation with the elite set size in the metaheuristic community, anecdotally, a hybrid tabu search with an elite set smaller than six performs much worse than larger elite sets on job shop scheduling problems.[2] In this paper, we experiment with elite set sizes of $\{1, 4, 8, 12, 16, 20\}$.

**The Proportion of Searches from an Empty Solution**  The $p$ parameter controls the probability of searching from an empty solution versus searching from one of the elite solutions. A high $p$ value will result in algorithm behaviour similar to randomized restart and indeed, $p = 1$ is a randomized restart algorithm. One reason that the $p$ parameter was included in SGMPCS was the intuition that it also has an impact on the diversity of the elite pool: the higher the $p$ value the more diverse the elite pool will be because solutions unrelated to the current elite solutions are more likely to enter the pool. As we will see in Experiment 2, this intuition is contradicted by our empirical results. Here, we study $p = \{0, 0.25, 0.5, 0.75, 1\}$.

**The Fail Limit Sequence**  The resource limit sets the number of fails allowed for each tree search. Rather than have a constant limit and be faced with the problem of tuning the limit (Gomes et al., 1998), following the work of Kautz, Horvitz, Ruan, Gomes, and Selman (2002), we adopt a dynamic restart policy where the limit on the number of fails changes during the problem solving. We look at two simple fail limit sequences (*seq*):

- Luby - the fail limit sequence is the optimal sequence for satisfaction problems under the condition of no knowledge about the solution distribution (Luby, Sinclair, & Zuckerman, 1993). The sequence is as follows: 1, 1, 2, 1, 1, 2, 4, 1, 1, 2, 1, 1, 2, 4, 8, .... That is, the fail limit for the first and second searches is 1 fail, for the third search is 2 fails, and so on. The sequence is independent of the outcome of the searches and of whether the search is from an empty solution or guided by an elite solution.

- Polynomial (Poly) - the fail limit is initialized to 32 and reset to 32 whenever a new best solution is found. Whenever a search fails to find a new best solution, the bound grows polynomially by adding 32 to the fail limit. The value 32 was chosen to give a reasonable increase in the fail limit on each iteration. No tuning was done to determine the value of 32. As with the Luby limit, the Poly fail limit is independent of the choice to search from an empty solution or from an elite solution.

---

2. Jean-Paul Watson – personal communication.

**Backtrack Method** Finally, as noted above, the style of an individual tree search is not limited to chronological backtracking. Whether search begins from an empty solution or an elite solution, we have a choice as to how the search should be performed. In particular, our backtracking ($bt$) factor is either standard chronological backtracking or limited discrepancy search (LDS) (Harvey, 1995). In either case, the search is limited by the fail limit as described above.

### 3.2 Experimental Details

Our experimental problems are job shop scheduling problem (JSP) instances. An $n \times m$ job shop scheduling problem consists of a set of $n$ jobs, each consisting of a complete ordering of $m$ activities. Each activity has a duration and a specified resource on which it must execute. The ordering of activities in a job represents a chain of precedence constraints: an activity cannot start until the preceding activity in the job has completed. Once an activity begins execution, it must execute for its complete duration (i.e., no pre-emption is allowed). There are $m$ unary capacity resources, meaning that each resource can be used by only one activity at a time. An optimal solution to a JSP is a sequence of the activities on each resource such that the union of the job sequences and resource sequences is acyclic, and the makespan (the time between the start of the earliest job and the end of the latest job) is minimized. The JSP is NP-hard (Garey & Johnson, 1979) and has received extensive study in both the operations research and the artificial intelligence literature (Jain & Meeran, 1999).

The experimental instances are twenty $20 \times 20$ problem instances generated using an existing generator (Watson, Barbulescu, Whitley, & Howe, 2002). The durations of the activities are independently drawn with uniform probability from [1, 99]. The machine routings are generated to create work-flow problems where each job visits the first 10 machines before any of the second 10 machines. Within the two machine sets, the routings are generated randomly with uniform probability. Work-flow JSPs are used as they have been shown to be more difficult than JSPs with random machine routings (Watson, 2003).

Each algorithm run has a 20 CPU minute time-out, and each problem instance is solved 10 times independently for a given parameter configuration. All algorithms are implemented in ILOG Scheduler 6.2 and run on a 2GHz Dual Core AMD Opteron 270 with 2Gb RAM running Red Hat Enterprise Linux 4.

For this experiment, the dependent variable is mean relative error (MRE) relative to the best solution known for the problem instance. The MRE is the arithmetic mean of the relative error over each run of each problem instance:

$$MRE(a, K, R) = \frac{1}{|R||K|} \sum_{r \in R} \sum_{k \in K} \frac{c(a, k, r) - c^*(k)}{c^*(k)} \quad (1)$$

where $K$ is a set of problem instances, $R$ is a set of independent runs with different random seeds, $c(a, k, r)$ is the lowest cost found by algorithm $a$ on instance $k$ in run $r$, and $c^*(k)$ is the lowest cost known for $k$. As these problem instances were generated for this experiment, the best-known solution was found either by the algorithms tested here or by variations used in preliminary experiments.[3]

---

3. Problem instances and best-known solutions are available from the author.

The variable ordering heuristic chooses a pair of activities on the same resource to sequence. Texture-based heuristics (Beck & Fox, 2000) are used to identify a resource and time point with maximum contention among the activities and to then choose a pair of unordered activities, branching on the two possible orders. The heuristic is randomized by specifying that the ⟨resource, time point⟩ pair is chosen with uniform probability from the top 10% most critical pairs. When starting search from an elite solution, the same heuristic is used to choose a pair of activities to be sequenced, and the ordering found in this solution is asserted. The standard constraint propagation techniques for scheduling (Nuijten, 1994; Laborie, 2003; Le Pape, 1994) are used for all algorithms.

### 3.3 Results

A fully crossed experimental design was implemented, consisting of four factors ($|e|$, $p$, $seq$, $bt$) with a total of 120 cells ($6 \times 5 \times 2 \times 2$). Each cell is the result of 10 runs of each of 20 problem instances, with a time limit on each run of 20 minutes. These results were generated in about 333 CPU days.

Analysis of variance (ANOVA) on the MRE at 1200 seconds shows that all factors and all interactions are significant at $p \leq 0.005$. The ANOVA is shown in Table 1.

| Factor(s) | Df | Sum Sq | Mean Sq | F value | Pr(>F) |
| --- | --- | --- | --- | --- | --- |
| e | 5 | 0.9995 | 0.1999 | 1229.6015 | < 2.2e-16 |
| p | 4 | 21.9376 | 5.4844 | 33736.2277 | < 2.2e-16 |
| bt | 1 | 0.8626 | 0.8626 | 5306.1350 | < 2.2e-16 |
| seq | 1 | 0.4924 | 0.4924 | 3028.6761 | < 2.2e-16 |
| e:p | 20 | 0.3735 | 0.0187 | 114.8711 | < 2.2e-16 |
| e:bt | 5 | 0.1144 | 0.0229 | 140.7780 | < 2.2e-16 |
| p:bt | 4 | 0.3023 | 0.0756 | 464.9442 | < 2.2e-16 |
| e:seq | 5 | 0.1359 | 0.0272 | 167.1942 | < 2.2e-16 |
| p:seq | 4 | 0.2265 | 0.0566 | 348.3872 | < 2.2e-16 |
| bt:seq | 1 | 0.0036 | 0.0036 | 22.1468 | 2.540e-06 |
| e:p:bt | 20 | 0.0372 | 0.0019 | 11.4361 | < 2.2e-16 |
| e:p:seq | 20 | 0.0503 | 0.0025 | 15.4859 | < 2.2e-16 |
| e:bt:seq | 5 | 0.0041 | 0.0008 | 5.0191 | 0.0001342 |
| p:bt:seq | 4 | 0.0078 | 0.0020 | 12.0281 | 9.144e-10 |
| e:p:bt:seq | 20 | 0.0105 | 0.0005 | 3.2147 | 1.547e-06 |
| Residuals | 23880 | 3.8821 | 0.0002 | | |

Table 1: Summary of the analysis of variance found using the R statistical package (R Development Core Team, 2006). All factors and all interactions are significant at $p \leq 0.005$.

To attain a more detailed view of the results, a Tukey HSD test (R Development Core Team, 2006) was performed for each of the factors. The Tukey HSD allows for the comparison of multiple means while controlling for the problems of multiple testing. Table 2 shows that, at significance level $p \leq 0.005$:

- Smaller $|e|$ is significantly better than larger $|e|$.

- $p = 0$ and $p = 0.25$ are not significantly different. However, they both result in significantly lower MRE than $p = 0.50$. For $p > 0.25$, a smaller value of $p$ is better.

- The Luby fail limit sequence is significantly better than the Poly sequence.

- Chronological backtracking is significantly better than LDS.

| | |
|---|---|
| $\|e\|$ | $1 < 4 < 8 < 12 < 16 < 20$ |
| $p$ | $\{0, 0.25\} < 0.50 < 0.75 < 1.00$ |
| $seq$ | Luby < Poly |
| $bt$ | chron < lds |

Table 2: The results of independent Tukey HSD tests on each factor. Significance level of the test on each parameter is $p \leq 0.005$. $a < b$ means that $a$ incurs a lower MRE than $b$, and the difference in MRE values is statistically significant. Parenthesis (i.e., {}) indicate no statistically significant difference in MRE.

Finally, Table 3 presents the five best and five worst parameter settings as determined by the MRE at 1200 CPU seconds. It is interesting to note that the five worst settings all have $p = 1.00$, which corresponds to a pure randomized restart algorithm.

| $\|e\|$ | $p$ | BT | Seq. | MRE |
|---|---|---|---|---|
| Five Best Parameter Settings | | | | |
| 1 | 0.25 | chron | Luby | 0.03158449 |
| 4 | 0.25 | chron | Luby | 0.03308468 |
| 1 | 0.25 | chron | Poly | 0.03328429 |
| 4 | 0.50 | chron | Luby | 0.03390888 |
| 1 | 0.50 | chron | Poly | 0.03421443 |
| Five Worst Parameter Setting | | | | |
| 4 | 1.00 | chron | Poly | 0.12637893 |
| 20 | 1.00 | chron | Poly | 0.12645527 |
| 1 | 1.00 | chron | Poly | 0.12651117 |
| 12 | 1.00 | chron | Poly | 0.12653876 |
| 8 | 1.00 | chron | Poly | 0.12711269 |

Table 3: The best and worst parameter combinations in Experiment 1 based on MRE.

A graphical representation of all results from this experiment is impractical. However, the statistical analysis is based on the performance of each set of parameter values at 1200 seconds, and so the evolution of performance over time is not reflected in these results. Given the arbitrariness of the 1200 second time limit, it is a valid question to wonder if the results would change given a different limit. To address this concern and to provide a graphical sense of the results, we present graphs of the experimental results where one parameter is varied and the others are held at their best values. For the parameters with only two values (i.e., $seq$ and $bt$) we display the results for two different values of $|e|$ as well.

**Elite Set Size:** $|e|$  Figure 1 shows the results of varying the elite set size with the other parameter settings as follows: $p = 0.25$, $seq = $ Luby, $bt = $ chron. The differences between the various levels of $|e|$ and the conclusion that lower $|e|$ results in better performance can be seen to hold for all time limits less than 1200 seconds. In fact, the superiority of the algorithms with small $|e|$ is most visible early in the search; after about 200 seconds, the gaps among the algorithms begin to narrow.

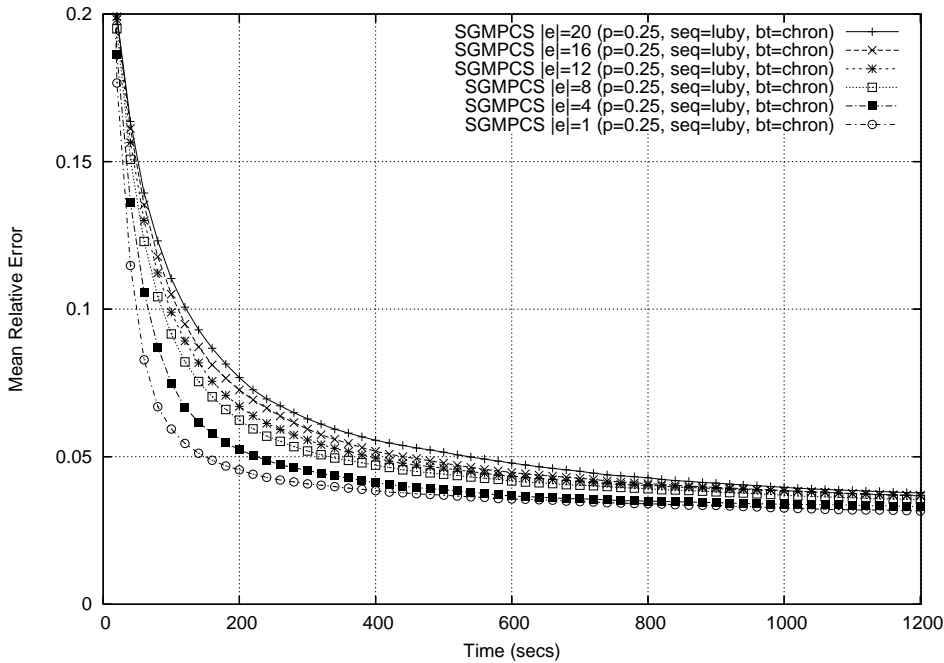

Figure 1: The mean relative error for SGMPCS on a set of makespan JSPs as the size of the elite set is varied.

Given the importance of diversity for elite solution sets within the metaheuristic literature, the performance of the algorithms with an elite size of 1 is somewhat surprising and seems to contradict some of our original intuitions and motivations for SGMPCS. We return to this point in Experiment 2.

**The Probability of Search from an Empty Solution:** $p$  Figure 2 displays the results of varying $p$ while holding the other parameter values constant at $|e| = 1$, $seq =$ Luby, and $bt =$ chron. The most dramatic result is the performance of $p = 1.00$, which is a pure randomized restart technique. All the other settings of $p$ result in performance that is more than an order of magnitude[4] better than $p = 1.00$.

Unlike in the experiments with the $|e|$ values, we do observe a change in the relative strengths of the different parameter settings with different time limits. While $p = 0.25$ results in the best performance for all time limits, for low limits $p = 0$ appears to out-perform $p = 0.50$ and $p = 0.75$. Later, the latter two parameter values result in better performance than $p = 0$. Note that this apparent contradiction of the statistical significance findings in

---

4. The MRE value achieved by $p = 1.00$ at 1200 seconds is achieved by all other $p$ values at less than 100 seconds.

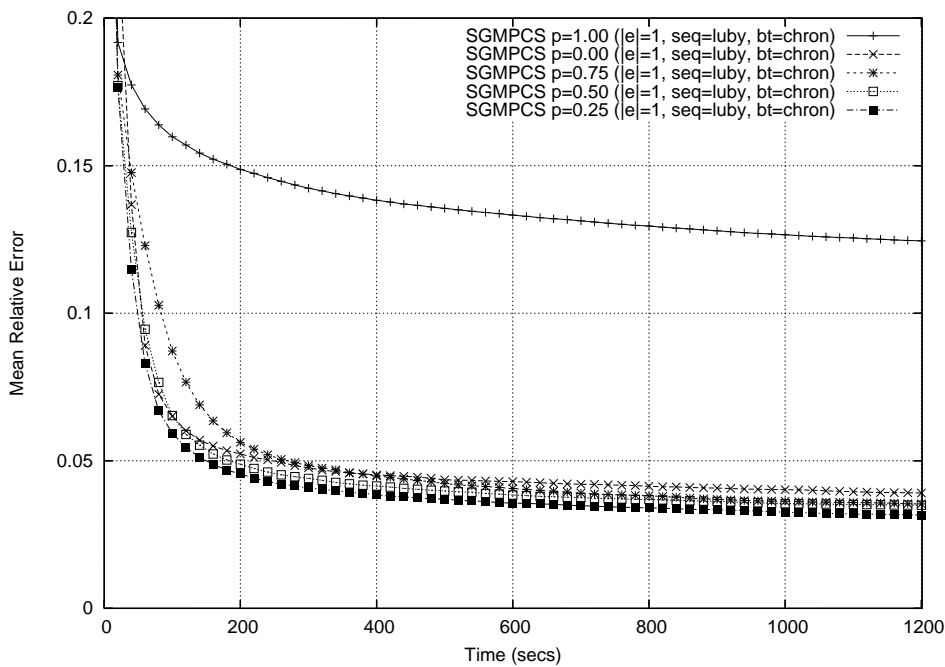

Figure 2: The mean relative error for varying $p$-values for SGMPCS on makespan JSPs.

Table 2 can be explained by the fact that there is interaction among the parameters and $p = 0$ performs better with other values of the rest of the parameters.

**Fail Sequence:** *seq*  Plots comparing the two different fail sequences are shown in Figure 3 with $p = 0.25$, $bt =$ chron, and with two different $|e|$ values, $|e| = 1$ and $|e| = 4$. For run-times less than about 100 CPU seconds, the Poly fail sequence performs better than the Luby sequence in both conditions. After that threshold, Luby performs better.

**Backtracking Method:** *bt*  Finally, Figure 4 displays the result of varying the backtracking method under the parameters of $p = 0.25$, $seq =$ luby, and $|e| = 1$ or $|e| = 4$. Using chronological backtracking for these problems clearly results in superior performance at all time limits when compared to LDS.

### 3.4 Summary

This experiment demonstrates that for job shop scheduling with makespan minimization, the best-performing parameter settings for SGMPCS are: a small elite set, a relatively low probability of starting search from an empty solution, the Luby fail limit sequence, and chronological backtracking. In general, these results are robust to changes in the time limit placed on the runs.

One should be careful in interpreting these results for a number of reasons.

1. As shown by the ANOVA, all the parameters have statistically significant interactions, and this was directly seen in the performance of $p = 0$, $|e| = 1$ in Figure 2.

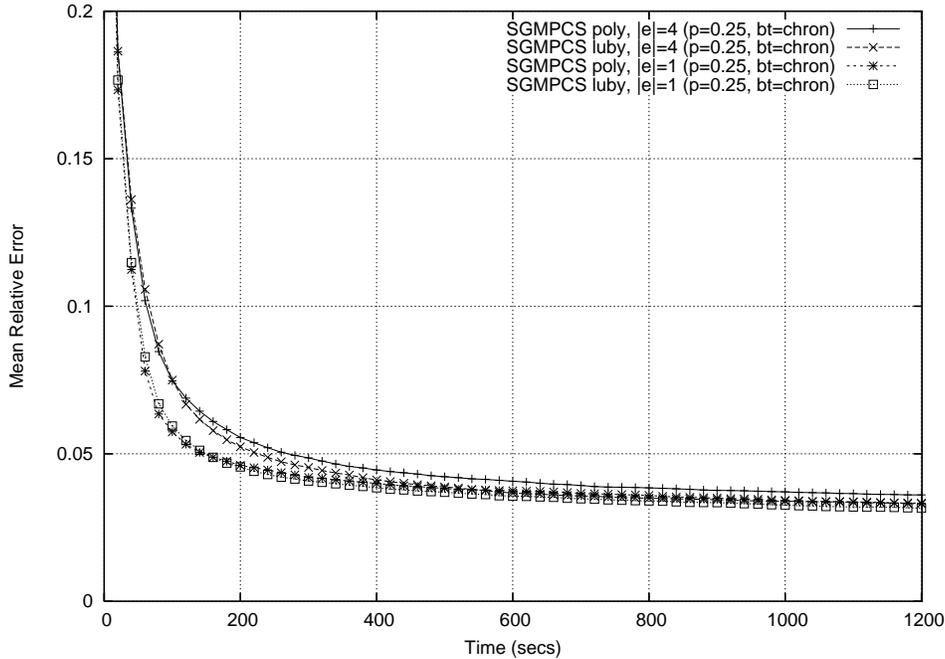

Figure 3: The mean relative error on makespan JSPs for the two different fail sequences for $|e| = 1$ and $|e| = 4$.

2. While there is a statistically significant effect for all factors, with the exception of the very poor performance of $p = 1.00$, the performance of different parameter settings displayed in the graphs are not wildly varying. While the differences among the levels of different factors may be statistically significant, they may not be practically significant. This is an advantage for SGMPCS as it suggests that fine tuning of parameters is not really necessary: SGMPCS is somewhat robust in the sense that small changes in parameters result in small changes in performance (again, with the exception of $p = 1.00$).

3. The results presented here are based on a single problem, job shop scheduling with makespan minimization. We comment on the applicability of these results to other problems in Section 6.2.

## 4. Experiment 2: The Impact of Elite Set Diversity

SGMPCS was designed with a number of intuitions about the impact of diversity on performance and on the likely effect of different parameter settings on performance. In particular, we test the following intuitions:

- A higher $|e|$ will tend to result in a higher diversity. This is not a strict relationship as it is possible that all solutions in $e$ are identical.

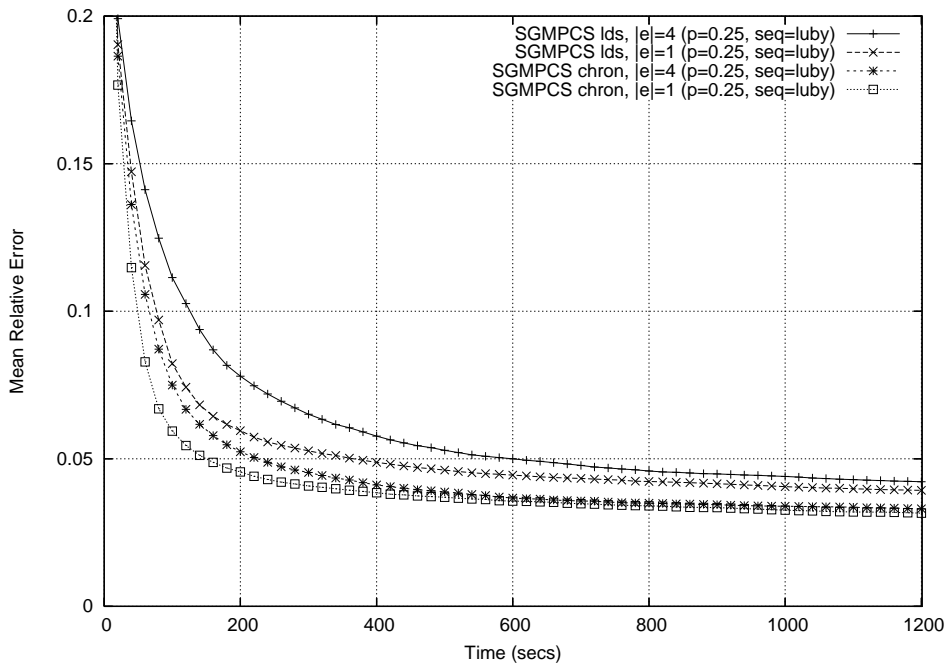

Figure 4: The mean relative error using the Luby fail limit and either chronological backtracking or LDS on the makespan JSPs.

- A higher $p$ value will tend to increase diversity. Since a higher $p$ increases the proportion of searches from an empty solution, it will lead to a wider exploration of the search space and therefore a more diverse elite set.

- The extent to which exploitation of multiple points in the search space is important for SGMPCS should be reflected in the performance of sets with different levels of diversity. That is, if it is important to simultaneously share search effort among a number of regions in the search space, we would expect that higher levels of diversity would out-perform lower levels up to some threshold of diminishing returns.

### 4.1 Measuring Diversity

The disjunctive graph (Pinedo, 2005) is a standard representation of a job shop scheduling problem where each activity is a node and the precedence constraints relating the activities in the same job are directed, *conjunctive* arcs. For each pair of activities in different jobs but on the same resource, there is a *disjunctive* arc: an arc that can be directed either way. In a solution, each disjunctive arc must be oriented in one direction so that the graph (which now contains only conjunctive arcs) is acyclic.

Following the work of Watson, Beck, Howe, and Whitley (2003), we measure the diversity of the elite pool by the mean pair-wise disjunctive graph distance. A binary variable is introduced for each disjunctive constraint where one value represents one orientation of the arc and the other value, the opposite orientation. A solution to the problem can therefore be represented by an assignment to these *disjunctive graph variables*. The distance between a pair of solutions is then simply the Hamming distance between the disjunctive graph variable assignments. For a given elite set, we take the mean pair-wise distance as a measure of diversity.

Clearly, this measure is not well-formed for $|e| = 1$. We assume that the diversity of an elite set of size 1 is 0.

### 4.2 Initial Evaluation of Diversity

Our initial evaluation of diversity is simply to measure the diversity for the problem instances and a subset of the parameter values used in Experiment 1. The SGMPCS solver was instrumented to calculated the pair-wise Hamming distance whenever a new solution was inserted into the elite set.

Figure 5 displays the diversity of the elite set over time for different elite set sizes. As expected, a higher elite set size results in a higher diversity. However, it is interesting to note the stability of the diversity: after the first 100 seconds, the diversity of the set changes very little, while the quality of solutions (see Figure 1) continues to improve.

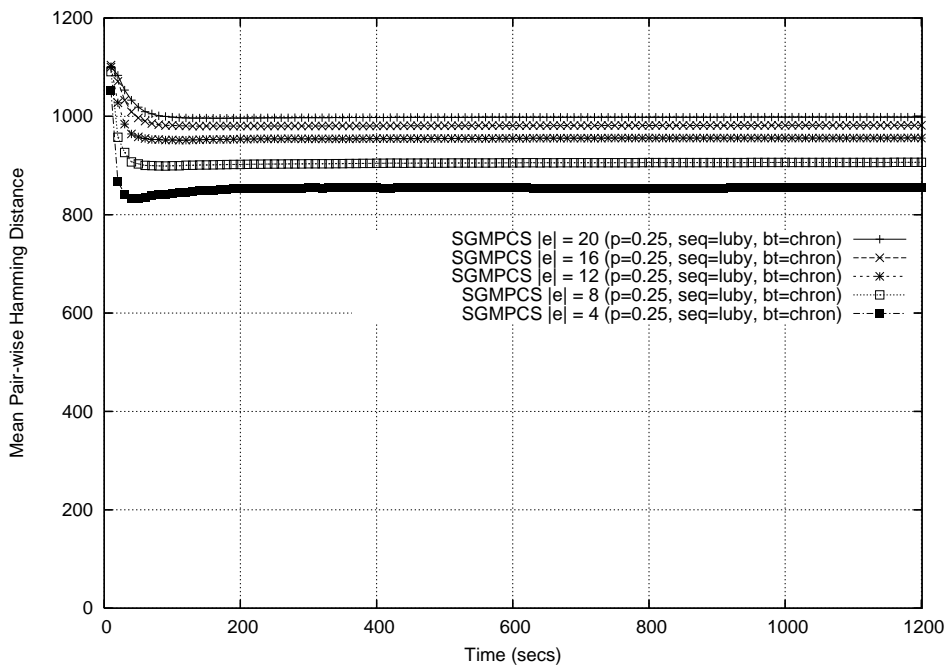

Figure 5: The diversity measured by mean pair-wise Hamming distance among the solutions in elite set for different elite set sizes.

Figure 6 shows the diversity with changing $p$ values. Contrary to our expectations, higher $p$ values exhibit *lower* diversity. Further analysis shows that the primary cause of this pattern is the way in which elite solutions are replaced. When search starts from an elite solution, an improved solution replaces the starting solution. Because the fail limit is relatively low, the starting solution is, with very high probability, also the closest elite solution to the improved solution. Therefore, replacing the starting elite solution has a relatively small impact on the overall diversity. In contrast, when search starts from an empty solution, the worst elite solution is replaced by an improved solution. As we demonstrate below, this difference in replacement policy results in a significantly lower elite pool diversity when more searches start from an empty solution: diversity decreases with increasing $p$.

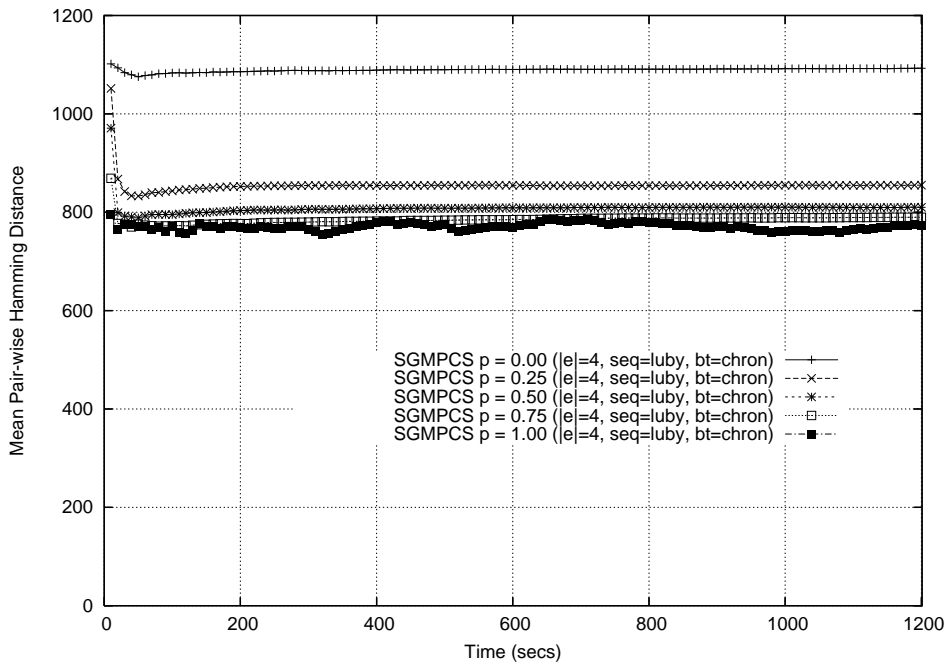

Figure 6: The diversity measured by mean pair-wise Hamming distance among the solutions in elite set for values of $p$.

### 4.3 Manipulating Diversity

Motivated by our interpretation of the results in Figure 6, in this section we experiment with the manipulation of the diversity by changing the elite solution replacement rule. Three levels of diversity are defined as follows:

- Low Diversity: Regardless of whether search starts from an elite solution or an empty solution, an improved solution replaces the worst elite one. No distance-based criteria is used. In the initialization phase, we follow the same approach as used above: 50 elite solutions are independently generated without constraining the makespan, and the $|e|$ best solutions inserted into the elite set.

- Medium Diversity: The standard elite set replacement rules used in Experiment 1 and defined in Section 2 are used.

- **High Diversity:** When search starts from an empty solution, the closest elite solution is replaced if an improving solution is found. When search starts from an elite solution, the starting solution replaced. As noted above, this latter rule results in the replacement of the closest solution with high probability. Therefore, these two rules are almost always equivalent to replacing the closest solution. During the initialization phase, $|e|$ solutions are generated and inserted into the elite pool. Then, an additional $50 - |e|$ solutions are generated and, if one of these solutions is better than the worst elite solution, the new solution is inserted into the elite set, replacing the closest elite solution.

To verify that our manipulations do indeed affect the diversity of the elite set as expected, we conduct an initial experiment over a subset of the parameter space. Using the problem instances from Experiment 1 and the same hardware and software configurations, we solved each problem instance 10 times under each diversity condition while varying $|e|$ and $p$. Rather than doing a fully crossed experiment, we set $|e| = 4$ and varied $p$ from 0 to 1, and set $p = 0.25$ and varied $|e|$ from 4 to 20.

Figures 7 and 8 demonstrate that the above manipulations affect the diversity of the elite set as expected. They show the different diversity levels with only two $|e|$ values and two $p$ values as displaying all the data was impractical. It is interesting to note that for the high and low diversity conditions, the effect on diversity of the other parameters disappears: there is little variation in the diversity when $|e|$ and $p$ are varied under those two diversity conditions.

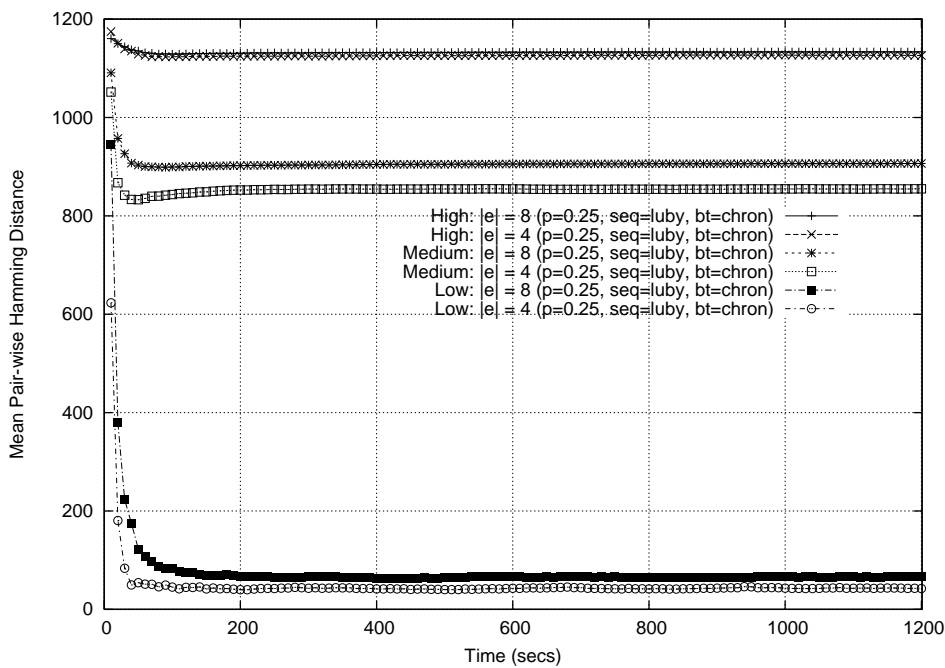

Figure 7: The diversity measured by mean pair-wise Hamming distance among the solutions in the elite set for different diversity levels for $|e| = 4$ and $|e| = 8$.

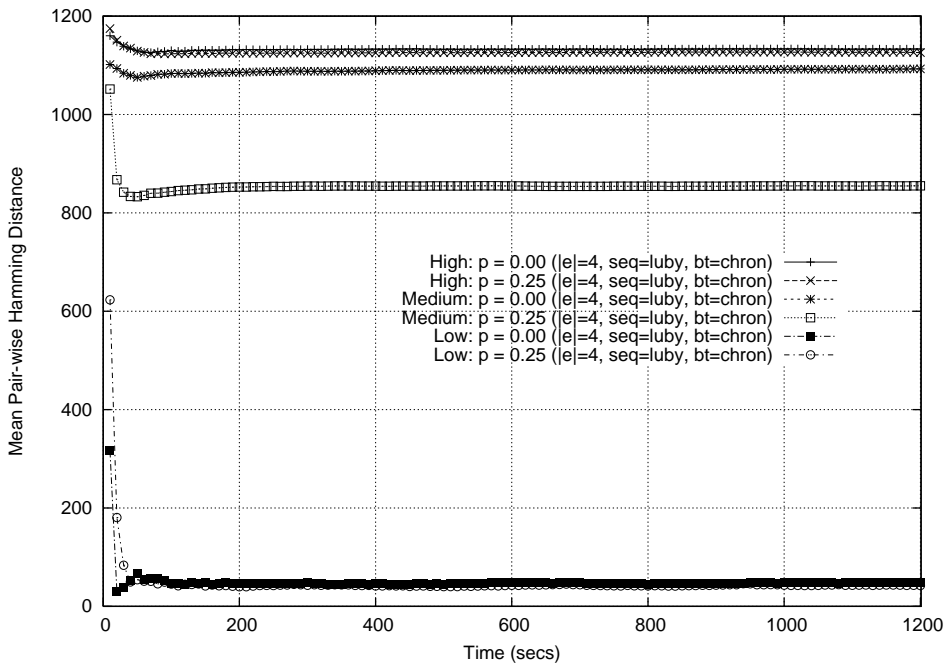

Figure 8: The diversity measured by mean pair-wise Hamming distance among the solutions in the elite set for different diversity levels for $p = 0$ and $p = 0.25$.

### 4.4 Experimental Details

Having verified that we do indeed have three different diversity settings, we can now test the impact of different diversity levels on the performance of SGMPCS. We perform a fully crossed experiment with three independent variables: $|e|$ which, as above, takes the values $\{1, 4, 8, 12, 16, 20\}$; $p$ which, as above, takes the values $\{0, 0.25, 0.5, 0.75, 1\}$; and diversity ($div$) taking the values low, medium, and high corresponding to the manipulations described above. In all conditions, we use chronological backtracking and the Luby fail limit sequence.

The other experimental details including the problem instances, hardware and software, the 1200 CPU second time limit, heuristics and propagators, and our evaluation criteria (MRE) are the same as in Experiment 1 (see Section 3.2).

### 4.5 Results

The fully crossed experimental design results in 90 cells ($6 \times 5 \times 3$). Each cell is the result of 10 runs of each of the 20 problem instances with a 20 minute time limit. These results were generated in about 250 CPU days.

The summary of the analysis of variance is shown in Table 4. These results demonstrate that all factors and all interactions are significant at $p \leq 0.005$. A Tukey HSD test (R Development Core Team, 2006) with significance level $p \leq 0.005$ was done on each of the factors, and the results are summarized in Table 5. The Tukey HSD results indicate that:

- As with Experiment 1, lower $|e|$ is better, though in this case there is no significant difference between $|e| = 1$ and $|e| = 4$.

- $p = 0$ is significantly worse than $p = 0.50$ which in turn is significantly worse than $p = 0.25$. Recall that in Experiment 1, $p = 0$ was not significantly different from $p = 0.25$.

- Lower diversity is better than medium which in turn is better than high diversity.

| Factor(s) | Df | Sum Sq | Mean Sq | F value | Pr(>F) |
|---|---|---|---|---|---|
| e | 5 | 0.0709 | 0.0142 | 81.9130 | < 2.2e-16 |
| p | 4 | 21.4690 | 5.3673 | 31000.5636 | < 2.2e-16 |
| div | 2 | 0.0706 | 0.0353 | 204.0232 | < 2.2e-16 |
| e:p | 20 | 0.0584 | 0.0029 | 16.8679 | < 2.2e-16 |
| e:div | 10 | 0.0234 | 0.0023 | 13.4938 | < 2.2e-16 |
| p:div | 8 | 0.0563 | 0.0070 | 40.6166 | < 2.2e-16 |
| e:p:div | 40 | 0.0186 | 0.0005 | 2.6925 | 4.298e-08 |
| Residuals | 17910 | 3.1008 | 0.0002 | | |

Table 4: Summary of the analysis of variance found using the R statistical package (R Development Core Team, 2006). All factors and all interactions are significant at $p \leq 0.005$.

| | |
|---|---|
| $|e|$ | $\{1, 4\} < 8 < \{12, 16\} < 20$ |
| $p$ | $0.25 < 0.50 < 0 < 0.75 < 1.00$ |
| div | low < medium < high |

Table 5: The results of independent Tukey tests on each factor in the diversity experiment. Significance level of the test on each parameter is $p \leq 0.005$.

Finally, Table 6 presents the parameter values that result in the five lowest and five highest MRE results. Note that the previous best set of parameter values ($|e| = 1$, $p = 0.25$, $div = med$) now incurs a slightly worse MRE than $|e| = 4$, $p = 0.25$, $div = low$.

### 4.6 Summary

Our experiment with diversity has addressed a number of our intuitions:

- As expected, a larger elite set results in a higher elite set diversity.

- Contrary to our expectations, a higher probability of searching from an empty solution *decreases* diversity. We were able to show that this impact was not directly due to the $p$ value but rather to the different elite set replacement rules.

- Finally, and most importantly, it appears that the diversity of the elite set is negatively correlated with performance: the lower the diversity, the higher the performance.

| $|e|$ | $p$ | Div | MRE |
|---|---|---|---|
| Five Best Parameter Settings | | | |
| 4 | 0.25 | low | 0.03085739 |
| 1 | 0.25 | med | 0.03158449 |
| 8 | 0.25 | low | 0.03224803 |
| 16 | 0.25 | low | 0.03231168 |
| 12 | 0.25 | low | 0.03233298 |
| Five Worst Parameter Setting | | | |
| 20 | 1.00 | med | 0.12482888 |
| 8 | 1.00 | low | 0.12484571 |
| 1 | 1.00 | low | 0.12487085 |
| 12 | 1.00 | med | 0.12488335 |
| 16 | 1.00 | low | 0.12489075 |

Table 6: Best and worst parameters for the diversity experiments.

This result calls into question the extent to which SGMPCS performance is based on exploiting multiple points in the search space. If such exploitation were important for performance, we would expect higher diversity to out-perform lower diversity. We return to this question in Section 6.

## 5. Experiment 3: Benchmark Comparison with Other Techniques

Our first two experiments concentrated on providing basic data on the performance of the different parameter settings of SGMPCS and an initial inquiry into the reasons underlying SGMPCS performance. In this experiment, we turn to comparisons of SGMPCS with existing heuristic search techniques.

### 5.1 Experimental Details

We use three sets of well-known JSP benchmark problem instances (Taillard, 1993). Each set contains 10 instances, and the different sets have problems of different size: $20 \times 15$, $20 \times 20$, and $30 \times 15$. The problems are numbered from 11 though 40.[5] These instances were not used during the development of SGMPCS.

Five algorithms are tested:

- Standard chronological backtracking (Chron): A non-randomized version of the same texture-based heuristic employed above is used together with the same global constraint propagators. As the heuristic is not randomized, one run is done for each problem instance.

- Limited Discrepancy Search (LDS): This is an identical algorithm to Chron except the backtracking is LDS.

---
5. See http://ina2.eivd.ch/collaborateurs/etd/problemes.dir/ordonnancement.dir/ordonnancement.html for the benchmark instances. The best-known upper and lower bounds are from the latest summary file on the same website, dated 23/11/05.

- Randomized Restart (Restart): This is a randomized restart algorithm using the same randomized texture-based heuristic and global constraint propagators used in Experiment 1 and 2. The backtracking between restarts is chronological and the fail limit used is the Luby limit. Each problem instance is solved 10 times.

- Solution-Guided Multi-Point Constructive Search (SGMPCS): We take the best parameters from Experiments 1 and 2: $|e| = 4$, $p = 0.25$, $seq = Luby$, $bt = chron$, and $div = low$. With these parameter settings, the sole difference between SGMPCS and Restart is the use of the elite set and the fact that some searches are guided by an elite solution. In particular, they use the same heuristics, propagators, fail limit sequence, and type of backtracking. Each problem instance is solved 10 times.

- Iterated Simple Tabu Search ($i$-STS): The $i$-STS algorithm is a sophisticated multi-phase tabu search built to model the state-of-the-art $i$-TSAB (Nowicki & Smutnicki, 2005) but with the goal of simplifying it in order to study how its various components contribute to the overall performance (Watson et al., 2006). On the Taillard benchmarks, $i$-STS only slightly under-performs $i$-TSAB in terms of solution quality given an equal number of iterations.[6] We use the parameters recommended[7] for the Taillard instances: $|E| = 8$, $X_a = 40000$, $X_b = 7000$, $p_i = p_d = 0.5$. For a full definition of these parameters, see the work of Watson et al. (2006).

The time limit for each run is 3600 CPU seconds. The other experimental details, including hardware and software for the first four algorithms, and the evaluation criteria are the same as in Experiment 1 (see Section 3.2). The $i$-STS algorithm is Watson et al.'s C++ implementation run on the same hardware as the other algorithms, meaning that direct run-time comparison is meaningful.

For all the constructive search-based approaches (i.e., all algorithms tested here except $i$-STS), the *Global Bound* policy is followed (see Section 2.2): whenever a new best solution is found, the global upper bound on the cost function is modified to be one less than this new best cost. In particular, this means that Restart benefits from the back-propagation of the cost constraint in exactly same way that SGMPCS does.

### 5.2 Results

The mean and best makespan found for each problem set are shown in Tables 7 through 9. Table 10 shows the performance in terms of finding and proving the optimal makespan for those problems for which the optimal solution is known.

#### 5.2.1 Comparing Constructive Search Algorithms

On the $20 \times 15$ problems (Table 7), SGMPCS dominates the other constructive algorithms, finding the lowest makespan (as judged by the mean makespan), for all but one instance (instance 14). In particular, on all problem instances the mean SGMPCS solution is better than the *best* solution found by Restart. In terms of mean relative error, SGMPCS outperforms each of the other constructive algorithms by a factor of about 3 to 8.

---

6. As with the previous experiments, here we use a CPU time limit. It is estimated that $i$-STS is about 5 to 7 times slower than $i$-TSAB.
7. Jean-Paul Watson, personal communication.

|       |           |       |       | Restart |      | SGMPCS |      | $i$-STS |      |
|-------|-----------|-------|-------|---------|------|--------|------|---------|------|
| Prob. | LB/UB     | Chron | LDS   | mean    | best | mean   | best | mean    | best |
| 11    | 1323/1359 | 1444  | 1410  | 1412.4  | 1408 | 1387.8 | **1365** | **1366.6** | **1365** |
| 12    | 1351/1367 | 1587  | 1411  | 1404.7  | 1402 | 1377.2 | **1367** | 1376.3 | 1375 |
| 13    | 1282/1342 | 1401  | 1401  | 1388.6  | 1385 | 1352.9 | **1343** | 1349.7 | 1347 |
| 14    | 1345      | 1496  | **1345**$^*$ | 1378.5 | 1370 | 1345.2 | **1345**$^*$ | 1345 | 1345 |
| 15    | 1304/1339 | 1436  | 1403  | 1432.2  | 1427 | 1375.9 | 1364 | **1350.2** | **1342** |
| 16    | 1302/1360 | 1496  | 1424  | 1416.2  | 1408 | 1373.3 | 1365 | **1362.3** | **1362** |
| 17    | 1462      | 1597  | 1485  | 1509.0  | 1507 | 1472.7 | **1462** | 1467.8 | 1464 |
| 18    | 1369/1396 | 1663  | 1464  | 1459.9  | 1456 | 1423.2 | **1400** | 1407.1 | 1404 |
| 19    | 1297/1335 | 1457  | 1388  | 1393.5  | 1386 | 1349.9 | **1335** | 1339.2 | **1335** |
| 20    | 1318/1348 | 1387  | 1390  | 1388.1  | 1378 | 1361.5 | 1356 | **1355.3** | **1350** |
| MRE (vs. UB) | | 0.0956 | 0.0343 | 0.0389 | 0.0348 | 0.0122 | 0.0036 | 0.0049 | 0.0026 |

Table 7: Results for Taillard's $20 \times 15$ instances. Bold entries indicate the best performance across the five algorithms on each instance. For Restart, SGMPCS, and $i$-STS, we use the mean makespan as the performance measure. We also include the best makespan found by the algorithms that solve an instance multiple times. The $^*$ indicates that the optimal makespan was found and proved for the problem instance. The final row shows the mean relative error (relative to the best-known upper bound) for each algorithm.

It is interesting to note the similar performance of LDS and Restart. We observe that when using a dynamic variable ordering, LDS performs "partial restarts" when jumping to the top of the tree to introduce a discrepancy. This suggests that some of the performance of LDS with dynamic variable orderings may be due to an exploitation of the heavy-tails phenomenon. The similar results here and on the other JSP instances in this section support this idea. To our knowledge this relationship has not been commented on before.

|       |           |       |       | Restart |      | SGMPCS |      | $i$-STS |      |
|-------|-----------|-------|-------|---------|------|--------|------|---------|------|
| Prob. | LB/UB     | Chron | LDS   | mean    | best | mean   | best | mean    | best |
| 21    | 1539/1644 | 1809  | 1699  | 1694.5  | 1686 | 1665.7 | 1649 | **1648.0** | **1647** |
| 22    | 1511/1600 | 1689  | 1659  | 1654.0  | 1649 | 1632.1 | 1621 | **1614.1** | **1600** |
| 23    | 1472/1557 | 1657  | 1620  | 1614.2  | 1602 | 1571.4 | 1561 | **1560.2** | **1557** |
| 24    | 1602/1646 | 1810  | 1676  | 1697.5  | 1694 | 1663.9 | 1652 | **1653.2** | **1647** |
| 25    | 1504/1595 | 1685  | 1669  | 1673.1  | 1664 | 1619.6 | 1608 | **1599.3** | **1595** |
| 26    | 1539/1645 | 1827  | 1723  | 1706.9  | 1701 | 1669.4 | 1656 | **1653.3** | **1651** |
| 27    | 1616/1680 | 1827  | 1755  | 1754.6  | 1750 | 1715.6 | 1706 | **1690.0** | **1687** |
| 28    | 1591/1603 | 1778  | 1645  | 1663.7  | 1656 | 1628.1 | 1619 | **1617.4** | **1614** |
| 29    | 1514/1625 | 1718  | 1678  | 1665.5  | 1660 | 1642.2 | **1626** | 1628.0 | 1627 |
| 30    | 1473/1584 | 1666  | 1659  | 1646.5  | 1641 | 1606.9 | 1598 | **1587.2** | **1584** |
| MRE (vs. UB) | | 0.0793 | 0.0373 | 0.0366 | 0.0324 | 0.0146 | 0.0072 | 0.0044 | 0.0019 |

Table 8: Results for Taillard's $20 \times 20$ instances. See the caption of Table 7.

Table 8 displays the results for the $20 \times 20$ problems. Again, SGMPCS dominates the other constructive algorithms, finding a mean makespan that is better than the *best* makespan found by any of the other constructive techniques. SGMPCS was unable to find solutions as good as the best-known upper bound for any of these instances. In terms of MRE, SGMPCS out-performs the other algorithms by a factor of 3 to 5.

| Prob. | LB/UB | Chron | LDS | Restart mean | Restart best | SGMPCS mean | SGMPCS best | $i$-STS mean | $i$-STS best |
|---|---|---|---|---|---|---|---|---|---|
| 31 | 1764 | 2118 | 1912 | 1896.8 | 1888 | 1774.0 | 1766 | **1764.0** | **1764** |
| 32 | 1774/1795 | 2163 | 1975 | 1983.1 | 1978 | 1828.3 | **1804** | 1813.4 | 1804 |
| 33 | 1778/1791 | 2138 | 1987 | 2021.6 | 2015 | 1840.9 | 1814 | **1804.2** | **1799** |
| 34 | 1828/1829 | 2096 | 1989 | 1968.4 | 1962 | 1863.9 | 1833 | **1831.9** | **1831** |
| 35 | **2007** | 2110 | **2007** | **2007.0** | **2007*** | **2007.0** | **2007*** | **2007.0** | **2007** |
| 36 | 1819 | 2411 | 1964 | 1957.1 | 1949 | 1832.7 | **1819*** | 1819.7 | 1819 |
| 37 | 1771 | 2018 | 1947 | 1940.3 | 1935 | 1810.6 | 1787 | **1791.1** | **1778** |
| 38 | 1673 | 2005 | 1853 | 1822.0 | 1817 | 1701.7 | 1691 | **1675.7** | **1673** |
| 39 | 1795 | 2118 | 1904 | 1896.1 | 1881 | 1803.5 | **1795*** | 1799.3 | 1797 |
| 40 | 1631/1674 | 2106 | 1870 | 1859.4 | 1855 | 1714.7 | 1690 | **1689.4** | **1686** |
| MRE (vs. UB) | | 0.190 | 0.0832 | 0.0813 | 0.0776 | 0.0147 | 0.0051 | 0.0044 | 0.0022 |

Table 9: Results for Taillard's $30 \times 15$ instances. See the caption of Table 7.

Table 9 displays the results on the largest problem instances ($30 \times 15$). On all instances but one, the mean solution found by SGMPCS is better than the best solution from each of the other constructive algorithms. For instance 35, SGMPCS equals the performance of LDS and Restart in finding (and, in some cases, proving) the optimal solution. Overall, SGMPCS is a factor of 5 to 13 better in terms of MRE.

| Prob. | Opt. | Chron | LDS | Restart | SGMPCS | $i$-STS |
|---|---|---|---|---|---|---|
| 14 | 1345 | 0(0) | **10(10)** | 0(0) | 9(9) | **10**(0) |
| 17 | 1462 | 0(0) | 0(0) | 0(0) | **1**(0) | 0(0) |
| 31 | 1764 | 0(0) | 0(0) | 0(0) | 0(0) | **10(10)** |
| 35 | 2007 | 0(0) | 10(0) | **10**(2) | **10**(4) | **10**(0) |
| 36 | 1819 | 0(0) | 0(0) | 0(0) | 1(**1**) | **8**(0) |
| 37 | 1771 | 0(0) | 0(0) | 0(0) | 0(0) | 0(0) |
| 38 | 1673 | 0(0) | 0(0) | 0(0) | 0(0) | **1(1)** |
| 39 | 1795 | 0(0) | 0(0) | 0(0) | **3(3)** | 0(0) |

Table 10: Results for the Taillard instances for which the optimal solution is known. The first two columns are the problem index and optimal makespan respectively. The rest of the columns are the number of runs for which each algorithm found an optimal solution and, in parenthesis, the number of times that it proved optimality. Recall that both Chron and LDS are run once per instance because they are not stochastic. However, to provide a fair basis of comparison, we present their results assuming they produced identical results in each of ten runs per instance. While $i$-STS is not a complete algorithm, there are some structural characteristics of a solution that imply optimality (Nowicki & Smutnicki, 1996). When a solution with such a characteristic is found, $i$-STS is able to prove optimality as shown in two instances: tai31 and tai38.

Finally, Table 10 presents the number of runs for which each algorithm was able to find and prove the optimal solutions for those problem instances with known optimal. SGMPCS finds the optimal solution at least once for five instances and proves the optimality at least once for four of those instances. Chron is unable to find or prove optimality for any instances, while Restart only does so for one instance, and LDS is able to find an optimal solution for two instances and prove it for one.

### 5.2.2 SGMPCS vs. $i$-STS

On almost all instances in Tables 8 and 9 $i$-STS performs substantially better than SGMPCS. In many cases, the mean solution found by $i$-STS is better than the best found by SGMPCS. However, on seven of the ten smallest instances (Table 7), the best solution found by SGMPCS is as good or better than that found by $i$-STS, and SGMPCS is strictly better on five instances. As with the larger problems, however, the mean makespan found by $i$-STS is better than that found by SGMPCS on all instances.

Recall that each algorithm was run for 3600 CPU seconds. While we do not include graphs of the run-time distributions, we have observed that the performance gap in terms of MRE between SGMPCS and $i$-STS at 3600 seconds is present at all time points from 60 seconds. In other words, $i$-STS substantially out-performs SGMPCS in the first 60 seconds and thereafter both algorithms find better solutions at about the same rate.

Table 10 shows that the one area that SGMPCS is clearly superior to $i$-STS is in proving the optimality of solutions. While $i$-STS is not a complete algorithm, it can identify solutions with a particular structure as optimal (Nowicki & Smutnicki, 1996). SGMPCS is able to find and prove optimality within the time limit on four instances in at least one run while $i$-STS can only do so for two instances.

### 5.3 Summary

Of the 30 problem instances used in this experiment, the mean solution found by SGMPCS was better than the *best* solution found by any of the other constructive techniques in 28 instances. Of the remaining instances, SGMPCS performs as well as LDS and Restart for instance 35 and slightly worse than LDS on instance 14. Overall, in terms of mean relative error, SGMPCS is between 3 and 13 times better than the other constructive search algorithms on the different problem sets.

SGMPCS does not perform as well as $i$-STS in terms of mean makespan; however, on the smaller problems the best solution it is able to find is better than that of $i$-STS on five instances.

## 6. Discussion and Future Work

This paper demonstrates that Solution-Guided Multi-Point Constructive Search can significantly out-perform existing constructive search techniques in solving hard combinatorial search problems but trails behind the state-of-the-art in metaheuristic search. In this section, we present some preliminary ideas regarding the reasons for the observed performance, a discussion of the generality of SGMPCS, and some directions for extensions of SGMPCS.

### 6.1 Why Does SGMPCS "Work"?

To the extent that SGMPCS out-performs existing constructive search approaches for solving hard combinatorial search problems, the most interesting question arising from the above experiments is understanding the reasons for this strong performance. We speculate that there are three, non-mutually exclusive, candidates: the exploitation of heavy-tails, the impact of revisiting previous high-quality solutions, and the use of multiple elite solutions.

#### 6.1.1 Exploiting Heavy-Tails

SGMPCS is a restart-based algorithm. Even with $p = 0$, search periodically restarts, albeit with a value ordering based on an elite solution. We believe that it is likely, therefore, that SGMPCS exploits heavy-tailed distributions in much the same way as randomized restart (Gomes et al., 2005; Gomes & Shmoys, 2002).

One way to test this idea is to reproduce Gomes et al.'s original experiment for SGMPCS as follows: for a random variable ordering, solve a problem instance to optimality starting from a given sub-optimal solution, $s$, and record the search effort involved; repeat for $k$ different random variable orderings for a large $k$; and finally observe the frequency distribution of search effort. The whole experiment can then be repeated for different starting solutions. If the resulting distributions exhibit heavy-tailed behaviour, the reasons that randomized restart is able to take advantage of heavy-tailed distributions may be shared by SGMPCS. We are currently pursuing such an experiment.

#### 6.1.2 Revisiting Solutions

While we believe it likely that the experiment suggested in Section 6.1.1 will demonstrate that SGMPCS takes advantage of heavy-tailed distributions, the significant performance advantage of SGMPCS over Restart in Experiment 3 as well as the very poor performance of the $p = 1$ parameter setting in Experiments 1 and 2, lead us to expect that there are additional factors needed to account for the performance of SGMPCS.

We believe that a leading candidate for one of these additional factors is the impact of revisiting high-quality solutions using a different variable ordering. Each time an elite solution is revisited with a different variable ordering, a different search tree is created. A resource-limited chronological search will only visit nodes deep in the tree before the resource limit is reached. However, a different variable ordering results in a different set of nodes that are deep in the tree and that are, therefore, within reach of the search.[8] The strong results of SGMPCS with $|e| = 1$ may be an indication that the mechanism responsible for the strong performance is the sampling of solutions "close" to an elite solution in different search trees. Our primary direction for future research is to formalize the meaning of "close" within a search tree to provide a firm empirical foundation on which to investigate the impact of revisiting solutions. We hope to adapt the significant work on fitness-distance correlation (Hoos & Stüzle, 2005) in the local search literature to constructive search.

---

8. Similar reasoning applies to the use of LDS.

### 6.1.3 Exploiting Multiple Points in the Search Space

The use of multiple solutions and, more specifically, the balance between intensification and diversification is viewed as very important in the metaheuristic literature (Rochat & Taillard, 1995). Intensification suggests searching in the region of good solutions while diversification suggests searching in areas that have not been searched before. Furthermore, one of the important aspects of the metaheuristics based on elite solutions is how the diversity of the elite set is maintained (Watson, 2005).

However, the experiments presented here suggest that increased diversity is not an important factor for performance of SGMPCS. The best performance was achieved with very small elite set sizes and even, in Experiment 1, an elite set size of 1. Based on such results, the original motivations for SGMPCS are, to say the least, suspect.

Our results may be due to idiosyncrasies of the makespan JSP problem. While experiments on some other problems (see below) have not directly manipulated diversity, the results have indicated better relative performance for larger elite set sizes than was observed here. This may be an indication that on other problems we will see a positive contribution of maintaining multiple viewpoints.

On a speculative note, a closer look at Figure 1 may show that diversity does play a role in search performance. That figure shows that the greatest differences in performance from different elite set sizes comes early in the search, where it is relatively easy to find an improving solution. Later in search, the performance difference narrows, though does not close completely within the time limit. One interpretation of this pattern is that, early in the search, when it is relatively easy to improve upon existing elite solutions, a large elite pool "distracts" the search by guiding it with an elite solution that is significantly worse than the best elite solution. The narrowing of the performance gap may be simply due to the fact that, with better solutions, it is harder to improve upon them and so regardless of the size of the elite set, the rate of improvement will decrease. Since the algorithms with lower $|e|$ have better solutions, their rate slows earlier. An alternative explanation is that maintaining multiple elite solutions has a positive influence only after the initial "easy" phase of search. When better solutions are harder to find, having a diverse elite set may help the search as the probability that at least one of the elite solutions has a better solution in its vicinity rises with the elite set size.[9] Further experimentation is required to investigate these intuitions.

### 6.2 Generality

SGMPCS is a general technique for conducting constructive search: nothing in the SGMPCS framework is specific to scheduling or constraint programming. However, in this paper only one type of problem was used to evaluate SGMPCS and therefore the question of its practical utility and generality should be addressed.

Existing work shows that SGMPCS can be effectively applied to other optimization and satisfaction problems such as quasigroup-with-holes completion (Beck, 2005b; Heckman & Beck, 2006), job shop scheduling with the objective to minimize weighted tardiness (Beck, 2006), and multi-dimensional knapsack optimization (Heckman & Beck, 2007). In addi-tion, recent work by Sellmann and Ansótegui (2006) demonstrates good performance of a closely related technique on diagonally ordered magic squares and some SAT instances. However, SGMPCS performs worse than randomized restart (though better than chronological backtracking) on magic square instances, and *both* randomized restart and SGMPCS perform much worse than chronological backtracking on a satisfaction version of the multi-dimensional knapsack problem (Heckman & Beck, 2006).

---

9. If this explanation is accurate, an adaptive strategy with $|e|$ growing during the search might be worth investigating.

The application of SGMPCS to such a variety of problems demonstrates that it is indeed a general technique whose impact can be applied beyond job shop scheduling. At the same time, the negative results on some problems point to our lack of understanding as to the mechanisms behind SGMPCS performance and motivates our future work.

### 6.3 Extending SGMPCS

While the immediate focus of our future work is on understanding the reasons for its performance, there are a number of ways in which the framework can be extended.

First, as implied by our speculations regarding the impact of diversity in Section 6.1.3, dynamic parameter learning (Horvitz et al., 2001) would appear very useful to the SGMPCS framework. For example, one could imagine adapting $p$ during the search depending on the relative success of searching from an empty solution versus searching from an elite solution.

Second, given that the metaheuristics community has been working with elite solutions for a number of years, there are a number of techniques which may fruitfully extend SGMPCS. For example, in path relinking (Glover, Laguna, & Marti, 2004) a pair of elite solutions is taken as end-points of a local search trajectory. Path relinking has an elegant counterpart in SGMPCS: two elite solutions are chosen, the variable assignments they have in common are fixed, defining a sub-space of the variable assignments in which the two solutions differ. Unlike in path relinking for local search, in constructive search one can perform a complete search of this sub-space and then post a no-good removing that sub-space from future consideration. Some preliminary experiments with such an approach appear promising.

Third, clause learning techniques, which originated as conflict learning in constraint programming (Prosser, 1993), are widely used with restart in state-of-the-art satisfiability solvers (Huang, 2007). It seems natural to investigate combining conflict learning and solution-guidance. These techniques may have an interesting relationship as the former tries to learn the "mistakes" that led to a dead-end while the latter attempts to heuristically identify the correct decisions that were made.

Finally, work on loosely coupled hybrid search techniques that share single solutions (Carchrae & Beck, 2005) is easily generalizable to share a set of solutions. To date, rather than being able to exploit a full solution shared by some other technique, constructive search is only able to use the bound on the cost function. Therefore, the revisiting of solutions provides a way to exploit the much richer information (i.e., full solutions) that is available in a hybrid search technique.

## 7. Conclusion

This paper presents the first fully crossed study of Solution-Guided Multi-Point Constructive Search. Using a set of job shop scheduling problems, we varied the SGMPCS parameter settings to control the size of the elite set, the probability of searching from an empty so-lution, the fail sequence, the form of backtracking, and the diversity level of the elite set. Experiments indicated that low elite set sizes, low probability of searching from an empty solution, the Luby fail sequence, chronological backtracking, and low diversity lead to the best performance. We then compared the best SGMPCS parameters found to existing constructive search techniques and to a state-of-the-art tabu search algorithm on a well-known set of benchmark problems. The results demonstrated that SGMPCS significantly outperforms chronological backtracking, limited discrepancy search, and randomized restart while being out-performed by the tabu search algorithm.

The primary contribution of this paper is the introduction of a new search framework and the demonstration that it can significantly out-perform existing constructive search techniques. Secondary contributions include the demonstration that the impact of elite set diversity on performance is the opposite of what was expected (i.e., low diversity leads to higher performance) and the identification of research directions into the reasons underlying the performance of SGMPCS by focusing on the quantification of the effects of heavy-tails, of the impact of revisiting solutions with different variable orderings, and of the exploitation of multiple points in the search space.

## Acknowledgments

This research was supported in part by the Natural Sciences and Engineering Research Council and ILOG, S.A. Thanks to Jean-Paul Watson, Daria Terekhov, Tom Carchrae, Ivan Heckman, and Lei Duan for comments on early versions of the paper. A preliminary version of parts of this work has been previously published (Beck, 2006).

## References

Beck, J. C. (2005a). Multi-point constructive search. In *Proceedings of the Eleventh International Conference on Principles and Practice of Constraint Programming (CP05)*, pp. 737–741.

Beck, J. C. (2005b). Multi-point constructive search: Extended remix. In *Proceedings of the CP2005 Workshop on Local Search Techniques for Constraint Satisfaction*, pp. 17–31.

Beck, J. C. (2006). An empirical study of multi-point constructive search for constraint-based scheduling. In *Proceedings of the Sixteenth International on Automated Planning and Scheduling (ICAPS'06)*, pp. 274–283.

Beck, J. C., & Fox, M. S. (2000). Dynamic problem structure analysis as a basis for constraint-directed scheduling heuristics. *Artificial Intelligence*, *117*(1), 31–81.

Carchrae, T., & Beck, J. C. (2005). Applying machine learning to low knowledge control of optimization algorithms. *Computational Intelligence*, *21*(4), 372–387.

Dilkina, B., Duan, L., & Havens, W. (2005). Extending systematic local search for job shop scheduling problems. In *Proceedings of Eleventh International Conference on Principles and Practice of Constraint Programming (CP05)*, pp. 762–766.

Garey, M. R., & Johnson, D. S. (1979). *Computers and Intractability: A Guide to the Theory of NP-Completeness*. W.H. Freeman and Company, New York.

Glover, F., Laguna, M., & Marti, R. (2004). Scatter search and path relinking: advances and applications. In Onwubolu, G., & Babu, B. (Eds.), *New Optimization Techniques in Engineering*. Springer.

Gomes, C. P., Selman, B., & Kautz, H. (1998). Boosting combinatorial search through randomization. In *Proceedings of the Fifteenth National Conference on Artificial Intelligence (AAAI-98)*, pp. 431–437.

Gomes, C. P., Fernández, C., Selman, B., & Bessière, C. (2005). Statistical regimes across constrainedness regions. *Constraints*, *10*(4), 317–337.

Gomes, C., & Shmoys, D. (2002). Completing quasigroups or latin squares: A structured graph coloring problem. In *Proceedings of the Computational Symposium on Graph Coloring and Generalizations*.

Harvey, W. D. (1995). *Nonsystematic backtracking search*. Ph.D. thesis, Department of Computer Science, Stanford University.

Heckman, I., & Beck, J. C. (2006). An empirical study of multi-point constructive search for constraint satisfaction. In *Proceedings of the Third International Workshop on Local Search Techniques in Constraint Satisfaction*.

Heckman, I., & Beck, J. C. (2007). An empirical study of multi-point constructive search for constraint satisfaction. *Submitted to Constraints*.

Hoos, H., & Stüzle, T. (2005). *Stochastic Local Search: Foundations and Applications*. Morgan Kaufmann.

Horvitz, E., Ruan, Y., Gomes, C., Kautz, H., Selman, B., & Chickering, M. (2001). A bayesian approach to tacking hard computational problems. In *Proceedings of the Seventeenth Conference on Uncertainty and Artificial Intelligence (UAI-2001)*, pp. 235–244.

Huang, J. (2007). The effect of restarts on the efficiency of clause learning. In *Proceedings of the Twentieth International Joint Conference on Artificial Intelligence (IJCAI07)*, pp. 2318–2323.

Hulubei, T., & O'Sullivan, B. (2006). The impact of search heuristics on heavy-tailed behaviour. *Constraints*, *11*(2–3), 159–178.

Jain, A. S., & Meeran, S. (1999). Deterministic job-shop scheduling: Past, present and future. *European Journal of Operational Research*, *113*(2), 390–434.

Jussien, N., & Lhomme, O. (2002). Local search with constraint propagation and conflict-based heuristics. *Artificial Intelligence*, *139*, 21–45.

Kautz, H., Horvitz, E., Ruan, Y., Gomes, C., & Selman, B. (2002). Dynamic restart policies. In *Proceedings of the Eighteenth National Conference on Artifiical Intelligence (AAAI-02)*, pp. 674–681.

Laborie, P. (2003). Algorithms for propagating resource constraints in AI planning and scheduling: Existing approaches and new results. *Artificial Intelligence*, *143*, 151–188.

Le Pape, C. (1994). Implementation of resource constraints in ILOG Schedule: A library for the development of constraint-based scheduling systems. *Intelligent Systems Engineering*, *3*(2), 55–66.

Luby, M., Sinclair, A., & Zuckerman, D. (1993). Optimal speedup of Las Vegas algorithms. *Information Processing Letters*, *47*, 173–180.

Nowicki, E., & Smutnicki, C. (1996). A fast taboo search algorithm for the job shop problem. *Management Science*, *42*(6), 797–813.

Nowicki, E., & Smutnicki, C. (2005). An advanced tabu algorithm for the job shop problem. *Journal of Scheduling*, *8*, 145–159.

Nuijten, W. P. M. (1994). *Time and resource constrained scheduling: a constraint satisfaction approach*. Ph.D. thesis, Department of Mathematics and Computing Science, Eindhoven University of Technology.

Pinedo, M. (2005). *Planning and Scheduling in Manufacturing and Services*. Springer.

Prestwich, S. (2002). Combining the scalability of local search with the pruning techniques of systematic search. *Annals of Operations Research*, *115*, 51–72.

Prosser, P. (1993). Hybrid algorithms for the constraint satisfaction problem. *Computational Intelligence*, *9*(3), 268–299.

R Development Core Team (2006). *R: A Language and Environment for Statistical Computing*. R Foundation for Statistical Computing, Vienna, Austria. ISBN 3-900051-07-0.

Rochat, Y., & Taillard, E. D. (1995). Probabilistic diversification and intensification in local search for vehicle routing. *Journal of Heuristics*, *1*, 147–167.

Sellmann, M., & Ansótegui, C. (2006). Disco-novo-gogo: Integrating local search and complete saerch with restarts. In *Proceedings of the Twenty-First National Conference on Artificial Intelligence (AAAI06)*, pp. 1051–1056.

Taillard, E. D. (1993). Benchmarks for basic scheduling problems. *European Journal of Operational Research*, *64*, 278–285.

Watson, J.-P. (2003). *Empirical Modeling and Analysis of Local Search Algorithms for the Job-Shop Scheduling Problem*. Ph.D. thesis, Dept. of Computer Science, Colorado State University.

Watson, J.-P. (2005). On metaheuristics "failure modes": A case study in tabu search for job-shop scheduling. In *Proceedings of the Fifth Metaheuristics International Conference*.

Watson, J.-P., Barbulescu, L., Whitley, L. D., & Howe, A. E. (2002). Contrasting structured and random permutation flow-shop scheduling problems: search-space topology and algorithm performance. *INFORMS Journal on Computing*, *14*(2), 98–123.

Watson, J.-P., Beck, J. C., Howe, A. E., & Whitley, L. D. (2003). Problem difficulty for tabu search in job-shop scheduling. *Artificial Intelligence*, *143*(2), 189–217.

Watson, J.-P., Howe, A. E., & Whitley, L. D. (2006). Deconstructing Nowicki and Smutnicki's *i*-TSAB tabu search algorithm for the job-shop scheduling problem. *Computers and Operations Research*, *33*(9), 2623–2644.